\documentclass[letterpaper]{article} 
\usepackage{aaai24}  
\usepackage{times}  
\usepackage{helvet}  
\usepackage{courier}  
\usepackage[hyphens]{url}  
\usepackage{graphicx} 
\usepackage{natbib}  
\usepackage{caption} 
\usepackage{algorithm}
\usepackage{algorithmic}
\usepackage{subcaption}
\usepackage{amsmath}
\usepackage{multirow}

\usepackage{newfloat}
\usepackage{listings}
\usepackage{tikz}
\def\checkmark{\tikz\fill[scale=0.4](0,.35) -- (.25,0) -- (1,.7) -- (.25,.15) -- cycle;} 
\DeclareCaptionStyle{ruled}{labelfont=normalfont,labelsep=colon,strut=off} 
\floatstyle{ruled}
\newfloat{listing}{tb}{lst}{}
\floatname{listing}{Listing}
 \nocopyright 

\title{Dynamic Y-KD: A Hybrid Approach to Continual Instance Segmentation}
\author{
    Mathieu Pagé Fortin, Brahim Chaib-draa\\
}
\affiliations{
    Laval University, Canada\\
    mathieu.page-fortin.1@ulaval.ca, brahim.chaib-draa@ift.ulaval.ca
}

\usepackage{bibentry}

\begin{document}

\maketitle

\begin{abstract}
Despite the success of deep learning models on instance segmentation, current methods {still} suffer from catastrophic forgetting in continual learning scenarios. In this paper, our contributions for continual instance segmentation are threefold. First, we propose the {\textit{Y-knowledge distillation} (\textit{Y-KD})}, a technique that shares a common feature extractor between the teacher and student networks. As the teacher is also updated with new data in {\textit{Y-KD}}, the increased plasticity results in new modules that are specialized on new classes. Second, our \textit{Y-KD} approach {is supported by} a dynamic architecture method that trains task-specific modules with a unique instance segmentation head, thereby significantly reducing forgetting. Third, we complete our approach by leveraging checkpoint averaging as a simple method to manually balance the trade-off between performance on the various sets of classes, thus increasing control over the model's behavior without any additional cost. These contributions are united in our model that we name the \textit{Dynamic Y-KD network}.
   
We perform extensive experiments on several single-step and multi-steps incremental learning scenarios, and we show that our approach outperforms previous methods both on past and new classes. For instance, compared to recent work, our method obtains $+2.1\%$ mAP on old classes in \textit{15-1}, $+7.6\%$ mAP on new classes in \textit{19-1} and reaches~$91.5\%$ of the mAP obtained by joint-training on all classes in \textit{15-5}.
\end{abstract}

\section{Introduction}
\begin{figure}[t]
    \centering
    \includegraphics[width=\linewidth]{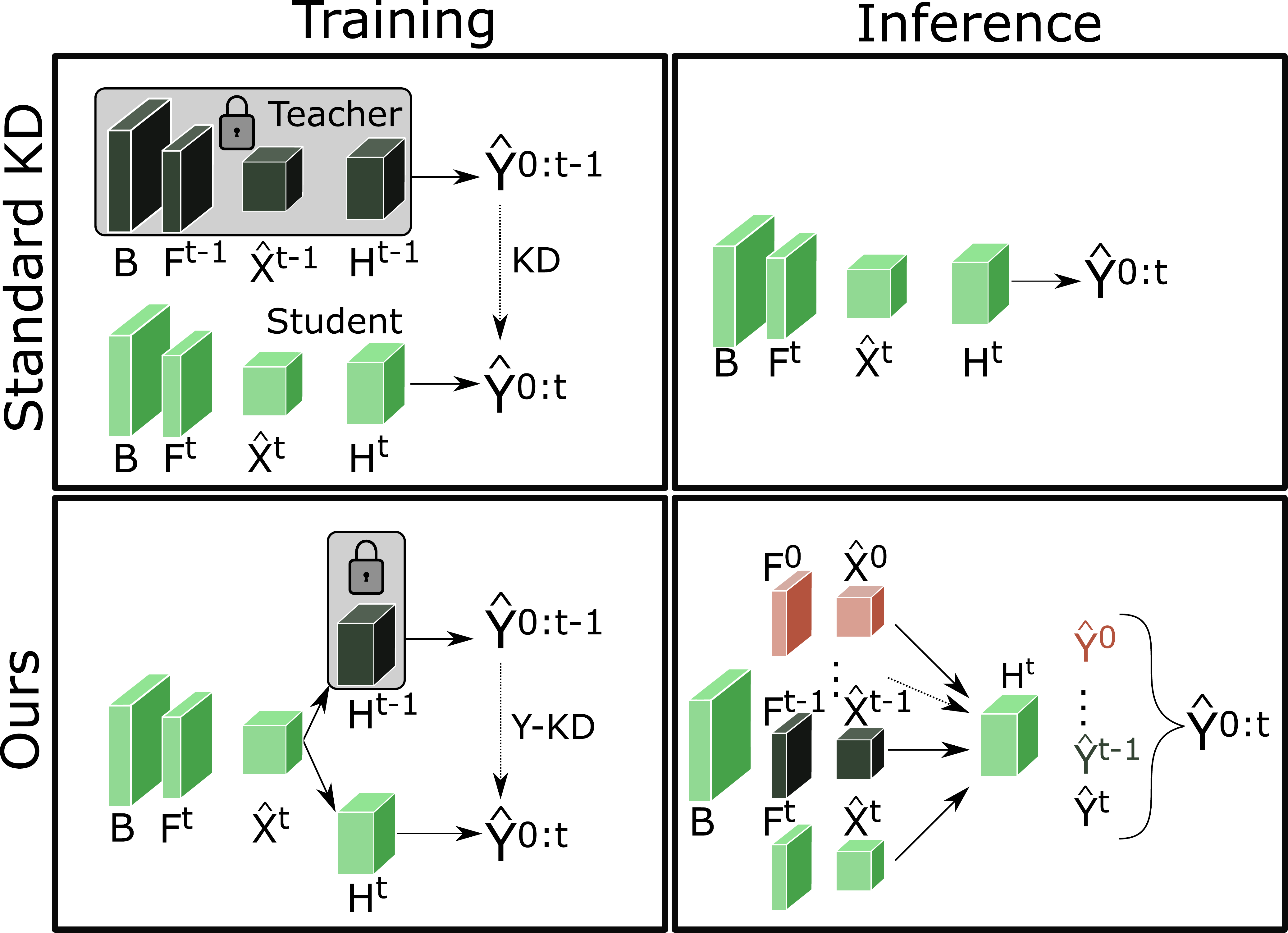}
    \caption{\small Overview of the differences between standard KD (top row) and our \textit{Dynamic Y-KD network} (bottom row). {During training, our approach only uses the previous head H$^{t-1}$ while the feature extractor is shared between the teacher and student networks. This increases plasticity, allowing to train task-specific feature extractors while preserving a generic head H$^t$. At inference, our \textit{Dynamic Y-KD network} benefits from more stability by using specialized modules in parallel with a generic instance segmentation head.}}
    \label{fig:overview}
\end{figure}

Instance segmentation, the task of detecting and segmenting each object individually in images, is a fundamental problem of computer vision that has many applications. Several approaches based on deep learning have been proposed in the last few years \cite{gu2022review}. However, it is generally assumed that the training dataset is fixed, such that training can be done in one step. This scenario faces limitations when deployed in real-world applications where the environment can change or use cases can evolve to include new sets of classes \cite{lesort2020continual}. Incrementing deep learning models to introduce new object categories is a challenging task because these methods are prone to catastrophic forgetting~\cite{mccloskey1989catastrophic}; they become biased towards novel classes while previous knowledge is discarded. 

{The challenge of catastrophic forgetting is encapsulated by the \textit{stability-plasticity dilemma} \cite{wu2021striking, grossberg1982studies}: a learning model must balance the preservation of past knowledge~(stability) with the flexibility to acquire new knowledge (plasticity). However, these two abilities generally conflict with one another. For instance, as gradient descent updates the weights of neural networks to learn new categories, this high plasticity also induces the replacement and suppression of previous knowledge \cite{de2021continual}.}


Continual learning (CL) is thus gaining more attention as it aims to bring deep learning methods to succeed even on non-stationary datasets. Previous work mostly studied CL for classification \cite{de2021continual}, and to a lesser extent semantic segmentation~\cite{cermelli2020modeling, douillard2021plop} and object detection \cite{menezes2023continual, wang2021wanderlust}. To our knowledge, the works of \citeauthor{gu2021class} \shortcite{gu2021class} and \citeauthor{cermelli2022modeling} \shortcite{cermelli2022modeling} are the only ones to propose continual instance segmentation (CIS) approaches. They both rely exclusively on knowledge distillation~(KD)~\cite{hinton2015distilling}, a popular regularization-based strategy that uses the model from the previous step as a teacher network to distill its knowledge in the new model, thereby reducing forgetting~(see Fig. \ref{fig:overview}, top row). However, the main drawback of KD is that performances are generally limited \cite{menezes2023continual}. KD constrains the model to increase stability but this comes at the cost of reduced plasticity, making new classes harder to learn optimally. 

{To address this, we propose in this paper a new} KD strategy in which the teacher and student networks share a common trainable feature extractor, coupled with a dynamic architecture model that grows new task-specific modules. {These two design choices are motivated by a preliminary study that highlights two key properties of Mask R-CNN trained in incremental scenarios, namely 1) the stability of feature extractors and 2) the compatibility of the head with previous feature extractors} (see Fig. \ref{fig:preliminary}). We name our hybrid approach the \textit{Dynamic Y-KD network}. 

Before learning new classes, the model from the previous step is duplicated and we only freeze the teacher instance segmentation head. Training images are then fed to the shared feature extractor and the resulting feature maps are sent in parallel 1)~to the new head for training and 2) to the previous head for KD, thus forming a Y-shaped architecture (see Fig. \ref{fig:overview}) that we name the \textit{Y-knowledge distillation} (\textit{Y-KD}). As the feature extractor of the teacher network is constantly updated, the student network benefits from more plasticity. This increased plasticity allows the growth of new feature extractor modules that are specialized on novel classes. 

During inference, the specialized modules are used with a unique instance segmentation head. Thus, by growing task-specific feature extraction branches to accommodate new categories, our model is able to learn new classes more efficiently, and by using specialized modules whose weights are frozen during incremental steps, forgetting of previous classes is significantly reduced. Notably, our results on Pascal-VOC~\cite{everingham2009pascal} and our ablation study show that the components of the \textit{Dynamic Y-KD network} enhance forward transfer \cite{menezes2023continual}. 

Moreover, if we measure performances of CL methods by their {mean average precision} (mAP) ratio with a non-CL equivalent (i.e. joint-training) \cite{menezes2023continual}, our approach obtains, on old classes, $\textbf{97.8}\%$ and $\textbf{89.0}\%$ compared to $94.7\%$ and $83.7\%$ obtained by MMA \cite{cermelli2022modeling} on \textit{19-1} and \textit{15-1}, respectively. On new classes in \textit{15-5} and \textit{10-2}, our approach obtains $\textbf{86.2}\%$ and $\textbf{83.1}\%$ of the joint-training mAP, compared to $78.9\%$ and $77.5\%$ obtained with MMA, respectively. 

Finally, inspired by our preliminary study that highlights the compatibility of incremented heads with previous feature extractors, we repurpose the use of checkpoint averaging~\cite{huang2017snapshot, gao2022revisiting} to provide control over the performance trade-off on different sets of classes in CL. Our results show that we can thereby easily adjust the model to either perform better on some sets of classes or others. This offers a simple control mechanism and can be a useful tool in the development of real-world applications where some classes are more important than others. 

In summary, our contributions are {as follows}:
\begin{itemize}
    \item {We highlight two intriguing properties of Mask R-CNN regarding 1) the stability of feature extractors, and 2) the compatibility of instance segmentation heads with previous feature extractors. To our knowledge, we are the first to make these observations.}
    \item {We exploit these two observations to propose 1) the \textit{Y-KD}, a new KD strategy that increases plasticity by using a shared feature extractor, and 2) a dynamic architecture that develops new task-specific feature extractors that are used with a common head at inference.} 
    \item Our \textit{Dynamic Y-KD network} significantly outperforms previous methods on various incremental scenarios of Pascal-VOC both on new and old classes. Furthermore, we isolate the contributions of each component in an ablation study. 
    \item We propose checkpoint averaging, a {zero-cost} mechanism to control the trade-off between performances on old, intermediary and new classes {after training}.
\end{itemize}

\section{Related Work}
\subsection{Instance Segmentation}
Instance segmentation is an important problem in computer vision that aims to produce a unique segmentation mask of objects that belong to a predefined set of classes. One of the most widely adopted approaches is the ``detect then segment'' strategy, which has been popularized by Mask R-CNN \cite{he2017mask}. 
Recent work on instance segmentation has explored alternative approaches such as one-stage methods \cite{bolya2019yolact, wang2020solov2}, and more complex techniques \cite{chen2019hybrid, fang2021instances, cheng2022masked}. However, few work addressed catastrophic forgetting when these methods face CL situations.

In this paper, we build upon Mask R-CNN as we propose a dynamic architecture that grows new modules of specialized feature extraction before the RPN to address the limitations of existing methods and improve the performance of instance segmentation in CL scenarios. 

\subsection{Continual Learning} 
CL studies solutions to enable the incrementation of models with novel classes without losing previously acquired knowledge. The main families of CL strategies are generally categorized into 1) replay-based \cite{rebuffi2017icarl, maracani2021recall, shieh2020continual, maracani2021recall, verwimp2021rehearsal}, 2) regularization-based \cite{cermelli2020modeling, cermelli2022modeling, liu2020incdet, kirkpatrick2017overcoming} and 3) dynamic architecture-based methods, also called parameter isolation-based \cite{rusu2016progressive, aljundi2017expert, li2018incremental, zhang2021incremental, douillard2022dytox}. In the following sections, we focus on regularization-based and dynamic architecture-based methods since we propose a hybrid strategy between these two approaches to build our \textit{Dynamic Y-KD network}.

\paragraph{Regularization-based Methods.} Since catastrophic forgetting results from a drift in the model's parameters, this can be mitigated by applying specific regularization losses. One of the most widely used regularization-based approaches is {knowledge distillation}~(KD) \cite{hinton2015distilling}, which leverages the outputs of a previous model to guide the new model in producing similar activations for previous categories. 

As examples, ILOD \cite{shmelkov2017incremental} applied a $L_2$ loss on the predicted logits of old classes and bounding boxes to prevent the new model from overly shifting its outputs towards new classes. In Faster ILOD \cite{peng2020faster}, an additional distillation term is applied on the features of the RPN of Faster-RCNN~\cite{girshick2015fast} for more stability. One of the first work on CIS has been proposed in \cite{gu2021class}, in which KD is performed by two teacher networks to increment YOLACT \cite{bolya2019yolact}. In MiB \cite{cermelli2020modeling}, the authors adapted the KD and cross-entropy losses to account for the background shift in continual semantic segmentation. In MMA \cite{cermelli2022modeling}, the authors then extended these ideas to the tasks of continual object detection and CIS with Faster R-CNN and Mask R-CNN respectively. 

In this work, we also leverage KD losses with Mask R-CNN. However, contrarily to previous work where the teacher network is completely frozen, our method differs as the feature extractor used for KD is shared with the learning model, and is therefore continuously updated during the learning process. This approach enhances the model's plasticity and forward transfer capabilities, as evidenced by our improved results on new classes and our ablation study (see Table \ref{tab:ablation} \textit{lines 2 vs 5}).  

\paragraph{Dynamic Architecture-based Methods.} These methods, also named parameter isolation, freeze some parts of the network \cite{li2018incremental} and grow new branches to learn new tasks~\cite{zhang2021incremental}. One of the drawbacks of this strategy is that it generally increases the memory footprint at each step. Some work such as \cite{zhang2021incremental} adopt model pruning to reduce the number of weights while limiting performance loss. In our work, we reduce model growth by showing empirically that a unique instance segmentation head can be used with small specialized feature extractors. Different strategies such as regularization and dynamic architectures each have their pros and cons. Our hybrid approach seeks to combine the strengths of both while mitigating their drawbacks. We thereby differ from previous work as we combine KD during training with a dynamic architecture approach to improve learning of new classes and reduce forgetting. 

\subsection{Checkpoint Averaging}
Averaging the weights from checkpoints saved at different epochs has been shown to improve generalization by acting similarly to ensemble methods \cite{huang2017snapshot, vaswani2017attention, gao2022revisiting}. In this work, we first show that this simple trick can also be leveraged in CL by averaging the weights between the instance segmentation heads trained after any incremental step $i$ and $j$ to reduce forgetting of classes $\mathcal{C}^{0:i}$ while preserving similar or slightly inferior results on new classes~$\mathcal{C}^j$. This offers a new mechanism to manually control the trade-off between performances on old and new classes without requiring retraining or incurring any additional cost.

\section{Continual Instance Segmentation}

\subsection{Problem Formulation}
In CIS, we aim to increment a model $f_{\theta^{t-1}}$, parameterized by $\theta^{t-1}$, to a model $f_{\theta^t}$ that can detect and segment instances of new classes $C^t$ as well as old classes $C^{0:t-1}$. At each step~$t$ we are given a training dataset $\mathcal{D}^t$ composed of images~$X^t$ and ground-truth annotations $Y^t$ that indicate the bounding boxes, segmentation masks and semantic classes. Following the experimental setup established in previous work \cite{cermelli2022modeling}, we consider that the annotations~$Y^t$ are only available for current classes $C^t$, whereas objects of previous categories {appearing} in $\mathcal{D}^t$ are unlabelled.

\subsection{Mask R-CNN for CIS}
In the context of CIS, Mask R-CNN \cite{he2017mask} is made of a feature extractor $F_{\theta^t}$ parameterized by $\theta^t$ at each step~$t$, a region proposal network (RPN) that proposes regions of interests (RoIs), and two parallel heads: 1) a box head for classification and regression of bounding boxes coordinates of each RoI, and 2) a segmentation head for the segmentation of each RoI. For simplicity, we summarize Mask R-CNN in three modules: 1) a backbone $B$ that is frozen during all steps, 2) a set of task-specific modules of feature extraction defined by $\{F_{\theta^{i}}\}_{i=0}^t$ that learn class-specific features from the outputs of $B$, and 3) a head $H_{\theta^t}$ that comprises the RPN, the box head and the segmentation head (see Fig. \ref{fig:overview}). 

\begin{figure*}
\centering
\begin{subfigure}{.4\textwidth}
    \centering
    \includegraphics[width=\linewidth]{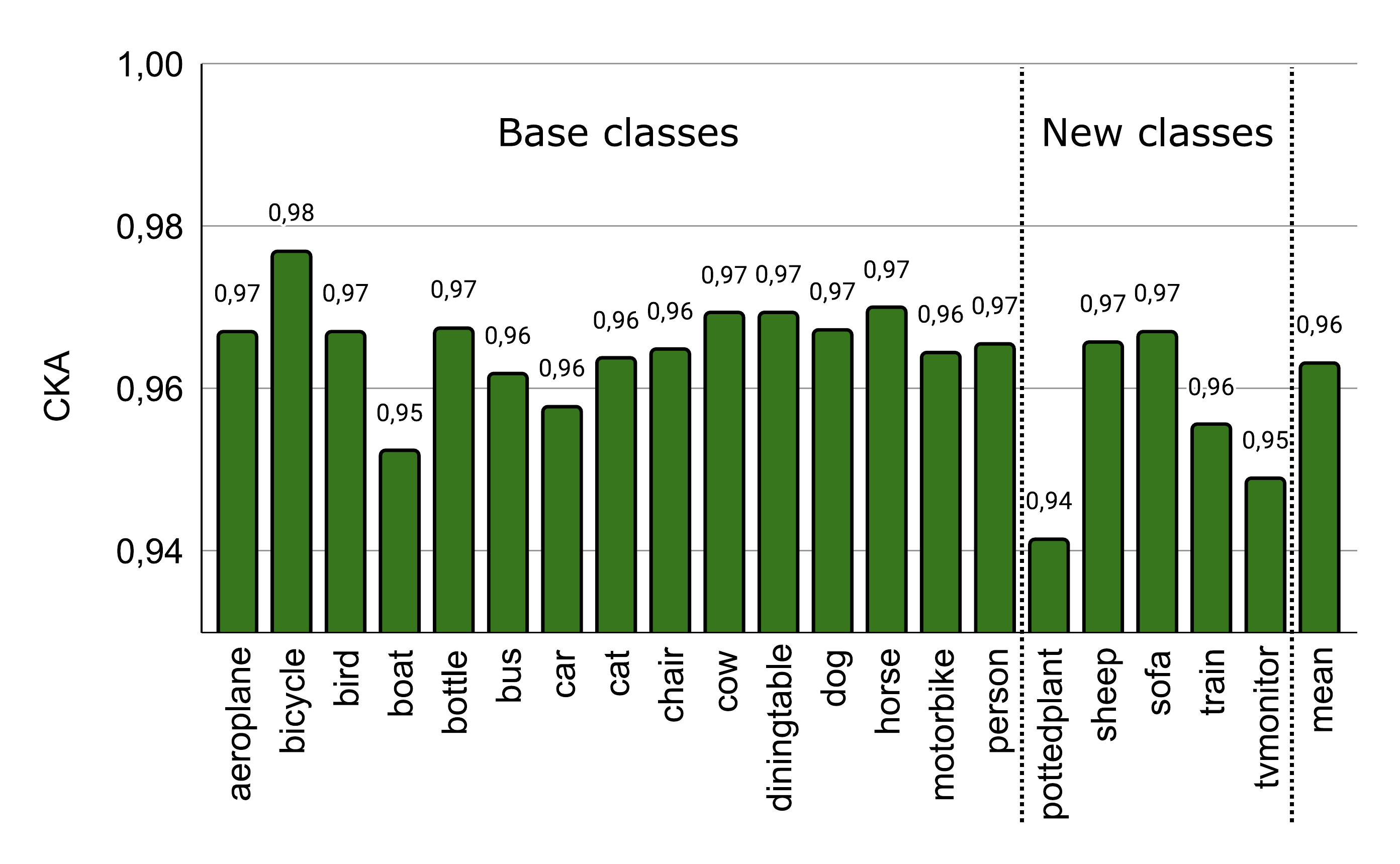}
    \caption{\small CKA betwen feature maps given by the backbone at~$t=0$ and $t=5$.}
    \label{fig:cka_classes}
    \end{subfigure} \quad\quad
\begin{subfigure}{.4\textwidth}
    \centering
    \includegraphics[width=\linewidth]{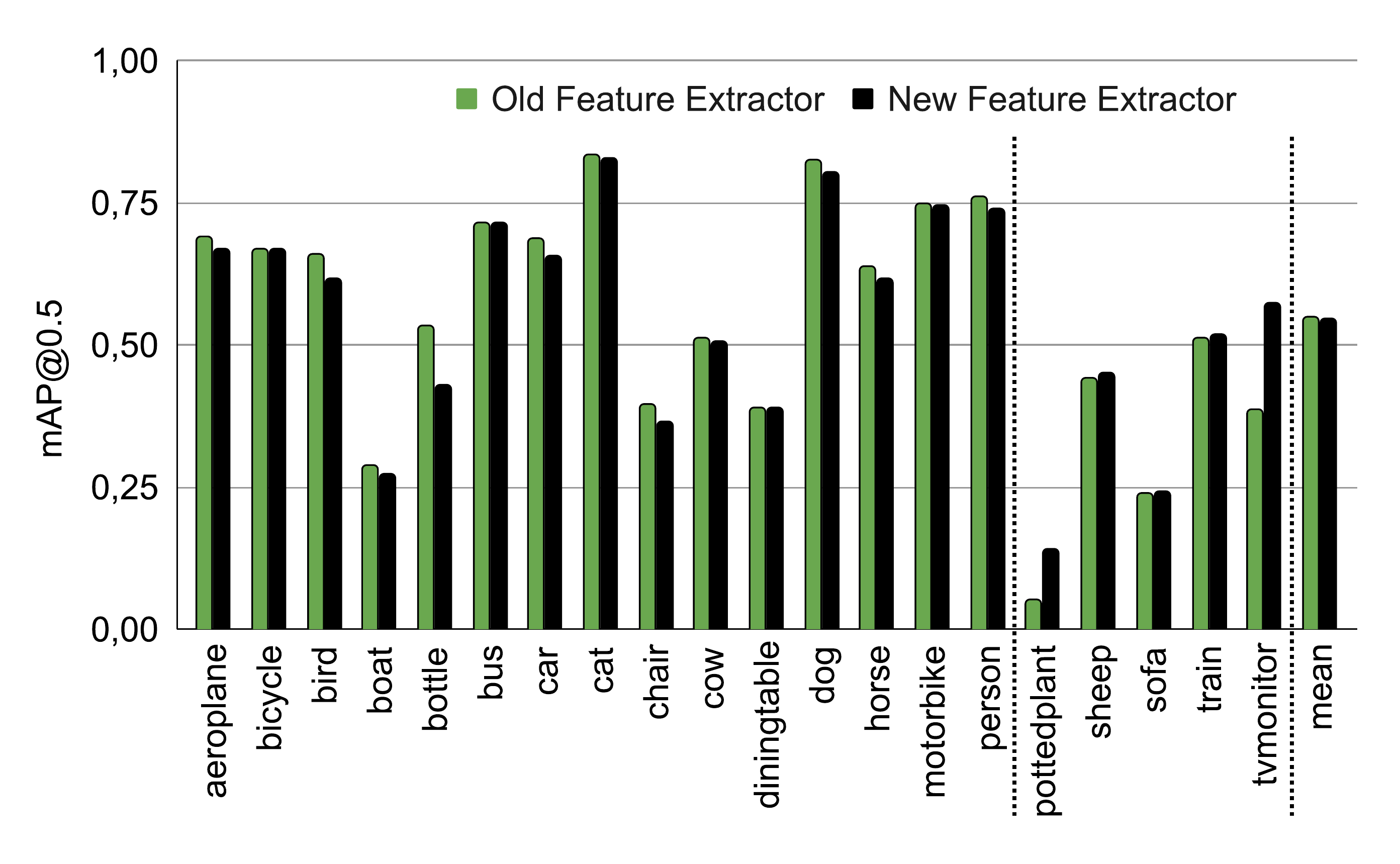}
    \caption{\small mAP@0.5 obtained with the new and old feature extactors, given the new instance segmentation head.}
    \label{fig:map_old_new}
\end{subfigure}
\caption{\small Preliminary experiments using Mask R-CNN with KD losses similar to \cite{cermelli2022modeling} in a \textit{15-1} scenario on Pascal-VOC.}
\label{fig:preliminary}
\end{figure*}

\paragraph{Knowledge Distillation.}
One of the main challenges of CL is to preserve past knowledge while learning new classes. Previous work \cite{shmelkov2017incremental, peng2020faster, cermelli2022modeling} showed the benefits of {knowledge distillation} (KD) to prevent the new network from significantly diverging while learning new classes. Generally, the KD loss has the following form:
\begin{equation}
    \mathcal{L}_{kd} = -\frac{1}{R\cdot C}\sum_{i=1}^{R}\sum_{c\in C^{1:t-1}}\hat{Y}_{i, c}^{t-1} \log \hat{Y}_{i, c}^{t},
\end{equation}
where $\hat{Y}_{i,c}^t$ is the score for class $c$ given by the model $f_{\theta^t}$ for the $i$-th output. In the context of RoI classification, this KD loss would encourage the new model to produce similar scores of past classes for each of the $N$ RoIs, i.e. $R~=~N$. On the other hand, since segmentation is a pixel-wise classification, the number of outputs is then $R=NH W$, where~$H$ and $W$ is the height and width of the segmentation masks, respectively. 

As highlighted in prior work \cite{cermelli2020modeling, cermelli2022modeling}, the conventional KD loss overlooks the background shift, wherein new classes were previously learned as background by the model. To address this, the KD loss should be adapted to incorporate the scores of these new classes into the background class before proceeding with distillation. The \textit{unbiased} KD loss \cite{cermelli2022modeling} thus becomes:

\begin{align*}
    \mathcal{L}_{unkd} = -\frac{1}{R\cdot C}&\sum_{i=1}^{R}  \Bigl[\hat{Y}_{i,bg}^{t-1}\log (\hat{Y}_{bg}^t + \sum_{c\in 
 C^t}\hat{Y}_{i,c}^t ) + \\
 & \sum_{c'\in C^{0:t-1} \backslash bg}\hat{Y}_{i, c'}^{t-1} \log \hat{Y}_{i, c'}^{t} \Bigl]. \tag{2}
\end{align*}

In this way, when the previous model gives high scores for the background class, the new model is encouraged to predict either \textit{background} or any of the new classes, which is the desired behaviour.

\section{Dynamic Y-KD: a Hybrid Approach}
In this section, we formulate our proposed \textit{Dynamic Y-KD network}. We begin by summarizing key observations that were made in preliminary experiments using Mask R-CNN with standard {knowledge distillation}~(KD) losses. From these observations, we motivate our \textit{Y-KD} and dynamic architecture strategies. We then proceed with a formulation of our hybrid method that synergistically leverages both techniques. Finally, we introduce checkpoint averaging as a mechanism to control the performance of CL models.

\subsection{Motivation}
\label{sec:motivation}
\paragraph{Stability of the Feature Extractor (FE).} In preliminary experiments on CIS using Mask R-CNN with standard KD losses, we noticed that the FE remains very stable even after several incremental steps. More specifically, we compared the representations produced by the base FE with the representations of the new FE that has been incremented with five novel classes after five incremental steps~(\textit{15-1}). We show in Figure \ref{fig:cka_classes} the Centered Kernel Alignment~(CKA) scores~\cite{kornblith2019similarity} of each class separately. Surprisingly, we found that the CKA scores were very high~(i.e.~$>0.94$), even for classes that have not been seen by the base model. This shows that the FE is only slightly fine-tuned to learn task-specific features during incremental steps. 

\paragraph{Compatibility of the head with previous FEs.} Then, we hypothesized that if the FE is stable, it should be possible to reuse the old FE with the new instance segmentation head. We compare in Figure \ref{fig:map_old_new} the mAP@0.5 of a model that uses either the new or the base FE with the same new head for inference. Interestingly, the base FE with the new head obtains better results on the old classes, showing the compatibility of the incremented head with a previous iteration of the FE. This highlights the compositionality of Mask R-CNN in CL: modules from different incremental steps are still compatible and can be effectively combined to give models with different properties. For instance, Figure \ref{fig:map_old_new} {shows} that using the FE from $t=0$ with the head at $t=5$ {produces} a model that is better on base classes but worst on new ones. This motivates the idea of developing task-specific FEs to preserve discriminative features of each set of classes. 

\subsection{Our Model}
\label{sec:dyn-ykd}

\paragraph{\textit{Y-KD}: Training Specialized Modules with a Generic Head.}
The stability of the FE suggests two aspects: 1) allowing the FE more plasticity may lead to improved results on new classes, and 2) it might not be necessary to freeze the teacher FE during KD if the FE is already stable.

We {have} explored these two hypotheses by proposing a KD strategy that aims to develop specialized FEs to better represent new classes while allowing the teacher FE to be updated. This is accomplished by using a common feature extractor $F_\theta^t$ which is connected in parallel to the previous head $H^{t-1}$ and the new head $H^t$, thus forming a Y-shaped architecture (see Fig. \ref{fig:overview}) that we name the \textit{Y-knowledge distillation}~(\textit{Y-KD}).    

In most previous works, a frozen copy of the whole teacher network is kept and used during training to distill its outputs to the student network. In our approach, \textit{Y-KD} consists of sharing the same trainable feature extractor between the teacher and student networks to increase plasticity during incremental learning. \textit{Y-KD} is thus performed by passing the images in the shared backbone and feature extractor, which gives the feature maps $\hat{X}^t$ as follows: 
\begin{align*}
    \hat{X}^t = F_\theta^t(B(X^t)). \tag{3}
\end{align*}
 The feature maps $\hat{X}^t$ are then sent to the teacher and student heads separately to produce their respective outputs:
\begin{align*}
    \hat{Y}^{t-1} &= H^{t-1}(\hat{X}^t),\\
    \hat{Y}^{t} &= H^{t}(\hat{X}^t), \tag{4}
\end{align*}
where $\hat{Y} := (p, r, s, \omega, m)$ which are respectively the class logits $p$, the regression scores $r$ of box coordinates, the objectness score $s$ and box coordinates $\omega$ given by the RPN, and the segmentation mask $m$. KD is then performed between the outputs of the teacher and student heads: 
\begin{align*}
    \mathcal{L}_{unkd}(\hat{Y}^{t-1}, \hat{Y}^{t}) =& \lambda_1 \mathcal{L}_{unkd}^{box}(p^{t-1},p^t, r^{t-1}, r^t) + \\
    &\lambda_2 \mathcal{L}_{kd}^{RPN}(s^{t-1}, s^t, \omega^{t-1}, \omega^t) +\\
    &\lambda_3 \mathcal{L}_{kd}^{mask}(m^{t-1}, m^t), \tag{5}
\end{align*}
where $\mathcal{L}_{unkd}^{box}$, $\mathcal{L}_{kd}^{RPN}$ and $\mathcal{L}_{kd}^{mask}$ are distillation losses \cite{cermelli2022modeling} applied on the {box head}, the RPN and the mask head, respectively. 

The total loss is then the following:
\begin{equation*}
    \mathcal{L} = \mathcal{L}_{mask}(\hat{Y}^t, Y^t) + \mathcal{L}_{unkd}(\hat{Y}^{t-1}, \hat{Y}^t), \tag{6}
\end{equation*}
where $\mathcal{L}_{mask}$ is the supervised loss to train Mask R-CNN. For more details on the specific implementation of these losses, we refer the reader to the supplementary material. 

With these distillation losses and by using a shared FE, the behaviour of the teacher network is {made} dynamic since its FE is also trained on novel images, but it still encourages the student head to preserve previous knowledge. This increases plasticity by allowing the student network to better learn the new classes while keeping the ability of the head to detect and segment previous categories.   

\paragraph{Dynamic Architecture.}
The second observation of our preliminary experiments, which highlighted the compatibility of the head with previous FEs, suggests that using task-specific FEs with a unique head would be a promising option for CIS. On the one hand, isolating parameters of task-specific FEs would reduce forgetting (as shown in Fig. \ref{fig:map_old_new}), and since a unique head would be used for inference, the growth in parameters would be {minimal}.

Therefore, we now propose our dynamic architecture-based method. At inference, we plug all specialized FEs to the same backbone and instance segmentation head in the following way (see Fig. \ref{fig:overview}). The backbone $B$ extracts general features from the input image $X$, and these features are sent to the task-specific modules $F^0, F^1, ... , F^t$ in parallel to produce their corresponding feature maps. These feature maps are then given to the most recent head $H^t$ for instance segmentation to produce their corresponding predictions~$\hat{Y}^0, \hat{Y}^1, ..., \hat{Y}^t$. All predictions are then merged by only keeping the outputs that correspond to the domain of expertise of each sub-network as follows:
\begin{equation*}
    \hat{Y}^t = [\hat{Y}_{c\in C^i}^i], \quad \forall i = 0, ... , t   \tag{7}
\end{equation*}

This filtering and merging step is necessary because we use a common generic head that can segment all classes from any feature maps.

\paragraph{Memory and Computational Costs.}
A common drawback of dynamic architecture-based strategies is that they generally increase the memory and computational costs as the model grows \cite{lesort2020continual}. Our approach does not make exception, as it linearly increases these costs by adding a specialized module for each task. However during training, we exclusively use the previous and new heads $H^{t-1}$ and $H^t$ with a shared backbone to perform our \textit{Y-KD}. Heads from earlier steps are discarded and previous specialized modules are not used during training (see Fig. \ref{fig:overview} bottom-left), such that the memory and computational costs are constant.

{Furthermore}, since a large part of the backbone is frozen, the number of weights added at each incremental step by the growth in task-specific FEs only accounts for $8.2M$ parameters, which represents $\frac{8.2M}{35.3M}=23.3\%$ of the original model when using ResNet-50. Future work should address this limitation, e.g., with pruning or quantization methods~\cite{zhang2021incremental}. In this paper, we focused on developing the first dynamic architecture for CIS.

\begin{table*}[t]
    \centering
    {
    \begin{tabular}{c|ccc|ccc|ccc}
    \hline
    \multicolumn{1}{c|}{}&\multicolumn{3}{c|}{\textbf{19-1}} & \multicolumn{3}{c|}{\textbf{15-5}}& \multicolumn{3}{c}{\textbf{10-10}}\\
    \textbf{Method} & \textbf{1-19} & \textbf{20} &\textbf{1-20} &\textbf{1-15} & \textbf{16-20} &\textbf{1-20} & \textbf{1-10} & \textbf{11-20} &\textbf{1-20}\\
    \hline
    Fine-tuning& 3.9 & 45.1 & 5.9 & 3.3 & 31.6 & 10.3&2.7&32.3&17.5\\
    \hline
    ILOD  &36.5 & 38.9 & 36.6 &  37.2 & 31.1 & 35.7 &36.1&25.9&31.0\\
    Faster ILOD  & 36.9 & 37.1 & 36.9 &  \textbf{37.7} & 30.7 & 35.9&37.0&25.8&31.4\\
    MMA  & 37.2 & 38.1 & 37.3 & 37.1 & 31.4 & 35.7 & 37.2 & 28.7&33.0\\
    Ours & \textbf{38.5} & \textbf{45.7} & \textbf{38.8} & {37.3} & \textbf{34.3} & \textbf{36.5}  & \textbf{37.4}&\textbf{29.8}&\textbf{33.6}\\
    \hline
    Joint Training& 39.3 & 50.0 & 39.9 & 39.9 & 39.8 & 39.9  & 40.0&39.7&39.9\\
    \hline
    \end{tabular}}
    \caption{\small mAP@(0.5, 0.95)\% results of single-step incremental instance segmentation on Pascal-VOC 2012.}
    \label{tab:benchmark-voc}
\end{table*}

\subsection{Checkpoint Averaging to Mitigate Forgetting}
In CL, the trade-off between performances on previous or new classes can only be indirectly controlled by choosing hyper-parameters before training~\cite{lesort2020continual, de2021continual}. 
However, this is a tedious approach as it requires retraining the model with different combinations until a satisfactory trade-off is reached. 

To alleviate this problem, we propose a simple tool to manually control the trade-off between performances on old and new classes by leveraging checkpoint averaging~\cite{huang2017snapshot, gao2022revisiting}. We can average the weights of heads that have been obtained after different incremental tasks to improve the ability of the model to segment instances of previous sets of classes. Given the parameters $\theta^i$ and $\theta^j$ of heads $H^i$ and $H^j$ that have learned classes $C^{0:i}$ and $C^{0:j}$ respectively, with $i<j$, we can create a new head $H_{\theta^m}^t$ that mixes their parameters as follows: 
\begin{equation*}
\label{eq:averaging}
    H_{\theta^m}^t := w_i \theta^i + w_j \theta^j, \tag{8}
\end{equation*}

where $w_i, w_j \in [0, 1]$ are factors to balance the contribution of each set of parameters. By doing so, we can recover performances on classes $C^{0:i}$ if forgetting is judged to be substantial. In return, a small drop in performance on classes $C^j$ should be expected. Nonetheless, this offers a simple zero-cost mechanism to gain control over forgetting, which can be a useful tool to define a performance balance, for instance when some classes are more critical then others.

\begin{table*}[t]
    \centering
    {
    \begin{tabular}{c|cccc|cccc|cccc}
    \hline
    \multicolumn{1}{c|}{}&\multicolumn{4}{c|}{\textbf{15-1}} & \multicolumn{4}{c|}{\textbf{10-2}}& \multicolumn{4}{c}{\textbf{10-5}}\\
    \textbf{Method} & \textbf{1-15} & \textbf{16-19} & \textbf{20}&\textbf{1-20} &\textbf{1-10} & \textbf{11-18}&\textbf{19-20} &\textbf{1-20} &\textbf{1-10} & \textbf{11-15} & \textbf{16-20}&\textbf{1-20}\\
    \hline
    Fine-tuning &-&-&-&-  & 0.5 & 0.4 & 40.1 & 4.4&1.2 & 0.5 & 31.4 & 8.6\\
    \hline
    ILOD &30.9 &19.7&39.9&29.1& 30.5& 17.6 & 39.6 & 26.2&35.9&25.8&29.2&31.7\\
    Faster ILOD &32.3&19.7&35.8&30.0& 30.5 & 17.8 & 38.5 & 26.2 &\textbf{36.1} & 26.1 & 29.1 & 31.9\\
    MMA  &33.4 & \textbf{21.2}&35.0&31.1 & 32.3 & \textbf{21.1} & 41.4 & 28.8 &  35.7 & \textbf{28.0} & 31.6 & \textbf{32.7}\\
    Ours & \textbf{35.5} & 19.7 & \textbf{43.9}&\textbf{32.8}& \textbf{33.5} & 20.0 & \textbf{44.4} & \textbf{29.2}& 35.7 & 25.2 & \textbf{33.4} & 32.5\\
    \hline
    Joint Training& 39.9&37.2&50.0&39.9& 40.0 & 36.3 & 53.4 & 39.9 &  40.0 & 39.7 & 39.8 & 39.9\\
    \hline
    \end{tabular}}
    \caption{\small mAP@(0.5, 0.95)\% results of multi-step incremental instance segmentation on Pascal-VOC 2012.}
    \label{tab:multi_steps}
\end{table*}
\section{Experiments}
\subsection{Experimental Setup}
Following previous work on CIS \cite{cermelli2022modeling, zhang2021incremental}, we opted to assess our approach using diverse continual learning scenarios derived from the Pascal-VOC dataset \cite{everingham2009pascal}. Given the increased complexity presented by the class-incremental scenario compared to conventional setups and the current state of continual instance segmentation methods, i.e. \cite{zhang2021incremental, cermelli2022modeling}, Pascal-VOC provides a more manageable benchmark than complex datasets that pose substantial challenges even in standard, non-continual learning contexts. 

Pascal-VOC is composed of $20$ semantic classes, which we divide in distinct sets to simulate incremental learning scenarios. Each scenario is defined as \textit{N-k}, where \textit{N} is the number of base classes in the first step, and \textit{k} is the number of classes added in the following incremental steps to reach the total of $20$ classes. 

\paragraph{Metrics.} We evaluate the performance of the models using the mean average precision (mAP), averaged over 10 thresholds ranging from 0.5 to 0.95, i.e., mAP@\{0.5:0.95\}. More specifically, we separately report 1) the mAP for base classes to show the ability to preserve past knowledge (i.e. stability); 2) the mAP for new classes to evaluate the capacity to be incremented with new categories (i.e. plasticity); 3) the mAP on all classes to show the global performance; and 4)~for multi-steps incremental learning, we also report the mAP of intermediary classes (e.g. classes \textit{16-19} in the \textit{15-1} scenario) separately since results on them are influenced both by plasticity and stability. In our analyses of the results, we also use the ratio between the mAP of a given CL method and its non-CL equivalent~(i.e. the joint training method) to give an idea of the level of performance that CL methods can achieve compared to a non-CL upper-bound.   

\paragraph{Baselines.} Since only very few methods have been proposed for CIS, we compare our approach with MMA \cite{cermelli2022modeling} as well as adaptations of ILOD \cite{shmelkov2017incremental} and Faster ILOD~\cite{peng2020faster} that have been presented in \cite{cermelli2022modeling}. We also consider lower and upper bounds, represented by a basic fine-tuning approach that does not incorporate any CL mechanism, and joint-training that trains on all classes simultaneously. We ran all experiments by extending the framework implemented by \cite{cermelli2022modeling}. 

\subsection{Results}

\paragraph{Single-step Incremental Learning.}
\label{sec:single-step}
The results for single-step incremental learning scenarios are shown in Table \ref{tab:benchmark-voc}. We can see that that the increased plasticity of fine-tuning, due to the absence of regularization losses, allows to obtain better results than most other methods on new classes. However, despite its superior plasticity, fine-tuning is far from achieving the same results than joint training on the new classes. 

On the other hand, our approach obtains significantly higher results on new classes while preserving similar or better mAP on base classes. On new classes, our approach obtains $+7.6\%$ in \textit{19-1}, $+2.9\%$ in \textit{15-5} and $+1.1\%$ in \textit{10-10} compared to MMA. Interestingly, our approach even outperforms fine-tuning on new classes in two of the three scenarios. This is especially the case in \textit{15-5} where we obtain $34.3\%$ on classes \textit{16-20} whereas fine-tuning, the second best approach, obtains $31.6\%$. This shows the ability of our \textit{Y-KD} strategy to enhance forward transfer by training feature extractors that are specialized on new classes. This contribution of \textit{Y-KD} is also highlighted by our ablation study below~(i.e. compare \textit{lines} 2-5 in Table \ref{tab:ablation})

In addition to giving better mAP on new classes, our approach also reduces forgetting compared to other methods, bringing the mAP on all classes (\textit{1-20}) closer to the ones of joint training in all three scenarios. Notably, compared to joint training, our method obtains mAP ratios of $\frac{38.8\%}{39.9\%}=97.2\%$ on classes \textit{1-20} in the \textit{19-1} scenario, $\frac{36.5\%}{39.9\%}=91.5\%$ in \textit{15-5}, and $\frac{33.6\%}{39.9\%}=84.2\%$ in \textit{10-10}.

\paragraph{Multi-steps Incremental Learning.}
\label{sec:multi-step}
We now show the results for multi-steps incremental learning scenarios in Table \ref{tab:multi_steps}. We can observe that our approach performs well in these more complicated situations. Our method stands out even more on base classes in the \textit{15-1} and \textit{10-2} scenarios, confirming the compatibility of previous FEs with an incremented head. Indeed, we outperform MMA by $+2.1\%$ and $+1.2\%$ on base classes in these scenarios, respectively. In \textit{10-5}, all approaches including ours obtain similar mAP ranging from $35.7-36.1\%$. On the last classes (e.g. classes \textit{16-20} in the \textit{10-5} scenario), our approach strongly outperforms previous work, as it obtains $+4.0\%$, $+3.0\%$ and $+1.8\%$ compared to the second best approaches in \textit{15-1}, \textit{10-2} and \textit{10-5}, respectively.

With respect to intermediary classes, the heightened plasticity in our method seems to come with a trade-off: it makes recently acquired knowledge more prone to forgetting. However, our method consistently outperforms others when evaluating performance across all classes (\textit{1-20}). For instance, even if our method obtains slightly inferior results on classes \textit{16-19} and \textit{11-18} in the \textit{15-1} and \textit{10-2} scenarios respectively, the \textit{Dynamic Y-KD network} obtains a better average on all classes (\textit{1-20}), outperforming MMA by $+1.7\%$ and $+0.4\%$. We now discuss how the checkpoint averaging trick can further address the limitation of our approach regarding intermediary classes.
\paragraph{Checkpoint Averaging.}
\label{sec:averaging}
\begin{table}[]
    \centering
    \begin{tabular}{c|cc|cccc}
    \hline
    Task&$w_4$&$w_5$&\textbf{Base}&\textbf{Int.}&\textbf{New}&\textbf{All}\\
    \hline
        \multirow{3}{*}{\textbf{15-1}} &0 &1 &35.5&19.7&\textbf{43.9}&32.8\\
        &0.25&0.75&{35.4}&\textbf{23.9}&42.7&\textbf{33.5}\\
        &0.5&0.5&\textbf{35.6}&{22.7}&39.7&{33.2}\\
        \hline
        \multirow{3}{*}{\textbf{10-2}}&0&1&33.5&20.0&\textbf{44.4}&29.2\\
        &0.25&0.75&33.8&20.2&42.9&\textbf{29.3}\\
        &0.5&0.5&\textbf{34.2}&\textbf{20.4}&40.0&29.2\\
    \hline
    \end{tabular}
    \caption{\small mAP@0.5-0.95\% results for checkpoint averaging using different weights $w_4$ and $w_5$.}
    \label{tab:my_label}
\end{table}

To ensure fairness, we did not use this trick while comparing methods in Tables \ref{tab:benchmark-voc}-\ref{tab:multi_steps}. We now show how our last contribution can be a viable tool to manage the compromise on different sets of classes, mitigating the drawback of our approach on intermediary classes. Specifically, we average the weights of the heads obtained after the fourth and fifth incremental learning steps according to Equation~\ref{eq:averaging}. The parameters $\theta_m$ of the new head used at inference thus becomes an average between the parameters of the heads $H_{\theta^4}$ and $H_{\theta^5}$, weighted by $w_4$ and $w_5$. 

In Table 3, we show the results on base, intermediary and new classes in \textit{15-1} and \textit{10-2} scenarios by varying the weights $w_4$ and $w_5$. We can see that although a small drop is observed on new classes, fusing the weights from the fourth incremental step allows to recover performances on intermediary and base classes. For instance, in \textit{15-1}, the decrease of mAP on new classes from $43.9\%$ to $42.7\%$ is compensated by an increase of $+4.2\%$ on intermediary classes~(i.e. \textit{16-19}) with $w_4=0.25$, which now outperforms previous methods (see Table \ref{tab:multi_steps}). Similarly, in \textit{10-2}, a slightly better mAP on all classes of $29.3\%$ can be obtained by using~$(w_4=0.25, w_5=0.75)$ as forgetting of base and intermediary classes is reduced by fusing past knowledge. Thereby, our proposed checkpoint averaging trick allows to manually create a new model that exhibits different performances on the various sets of classes without requiring any training or additional computational costs.
\paragraph{Ablation Study.}
\begin{table}[]
    \centering
    \begin{tabular}{ccccc|ccc}
    \hline
       &\textbf{Y-KD}  & \textbf{KD} & \textbf{FE$^0$} &\textbf{FE$^1$}&\textbf{1-15}&\textbf{16-20}&\textbf{1-20} \\
    \hline
         1&&&\checkmark&\checkmark&21.7&52.9&29.5\\
         2&&\checkmark&\checkmark&\checkmark&67.2&53.1&63.7\\
         3&\checkmark&&&\checkmark&60.8&\textbf{57.8}&60.0\\

        4&\checkmark&&\checkmark&&\textbf{67.4}&37.2&59.8\\
        5& \checkmark&&\checkmark&\checkmark&\textbf{67.4}&\textbf{57.8}&\textbf{65.0}\\
         \hline
    \end{tabular}
    \caption{\small mAP@0.5\% results of the ablation study in the \textit{15-5} scenario.}
    \label{tab:ablation}
\end{table}
\label{sec:ablation}
Finally, we perform an ablation study to highlight the importance of the two aspects of our \textit{Dynamic Y-KD network}, namely that 1) \textit{Y-KD} using a shared FE during training improves results on new classes and 2) using a dynamic architecture reduces forgetting.

The results of the ablation study in a \textit{15-5} scenario are shown in Table \ref{tab:ablation}. From \textit{line 1}, we can see that a purely architectural-based strategy that grows new modules to accommodate new tasks does not work for CIS, as catastrophic forgetting still happens. Without KD, FE$^0$ cannot remain compatible with the incremented head, such that it performs poorly on previous classes. 
While standard KD (\textit{line 2}) offers reasonable performances on old and new classes, we can see that new classes can be better learned using our \textit{Y-KD} strategy (\textit{line 3}). However, the increased plasticity from using a shared backbone in our \textit{Y-KD} strategy comes at a cost of decreased stability, as shown by the fact that the mAP@0.5 drops to $60.8\%$ on classes \textit{1-15}. Better results on these previous classes can be obtained using FE$^0$ (\textit{line 4}), as the mAP@0.5 rises to $67.4\%$. But since FE$^0$ has not learned task-specific features of classes \textit{16-20}, it cannot perform as well on new classes. Therefore, the best of both worlds is obtained by using both FE$^0$ and FE$^1$ (\textit{line 5}), which corresponds to our \textit{Dynamic Y-KD network}, as it performs better on new classes $(\textbf{57.8\%}$ \textit{vs} $53.1\%$) and even slightly better on previous classes than standard KD ($\textbf{67.4\%}$ \textit{vs} $67.2\%$).

\section{Conclusion}
In preliminary experiments on continual instance segmentation using Mask R-CNN with knowledge distillation, we made two observations regarding the stability of feature extractors and the compatibility of instance segmentation heads with previous backbones. We leveraged these two observations by proposing the \textit{Y-KD} and the use of a dynamic architecture to form the \textit{Dynamic Y-KD network}. Our approach increases plasticity and allows to train feature extractors that are specialized on new classes, while preserving a generic head that is compatible with all previous task-specific feature extractors for better stability.

Our results on several single-step and multi-steps incremental learning scenarios showed that our approach reduces forgetting of previous classes as well as improving mAP on new classes, thus outperforming previous methods in most setups. Additionally, we proposed a zero-cost trick based on checkpoint averaging to manually adjust the trade-off between the performances on the various sets of classes.

\bibliography{paper}

\end{document}